\documentclass[runningheads]{readmission}

\usepackage{csquotes}
\usepackage{amssymb}
\usepackage{lscape}
\usepackage{graphicx}
\usepackage{caption}
\usepackage{subcaption}
\usepackage{booktabs}
\usepackage{multirow}
\usepackage{xspace}
\usepackage{siunitx}
\usepackage{booktabs}
\usepackage{array}
\usepackage{enumitem}
\usepackage[bottom]{footmisc}

\makeatletter
\def\set@curr@file#1{%
  \begingroup
    \escapechar\m@ne
    \xdef\@curr@file{\expandafter\string\csname #1\endcsname}%
  \endgroup
}
\def\quote@name#1{"\quote@@name#1\@gobble""}
\def\quote@@name#1"{#1\quote@@name}
\def\unquote@name#1{\quote@@name#1\@gobble"}
\makeatother

\setlist{parsep=0pt,listparindent=\parindent}

\emergencystretch=\maxdimen
\newcolumntype{L}[1]{>{\raggedright\let\newline\\\arraybackslash\hspace{0pt}}m{#1}}
\newcolumntype{C}[1]{>{\centering\let\newline\\\arraybackslash\hspace{0pt}}m{#1}}
\newcolumntype{R}[1]{>{\raggedleft\let\newline\\\arraybackslash\hspace{0pt}}m{#1}}

\newsavebox\IBoxA \newsavebox\IBoxB \newlength\IHeight
\newcommand\TwoFig[6]{
  \sbox\IBoxA{\includegraphics[width=0.7\textwidth]{#1}}
  \sbox\IBoxB{\includegraphics[width=0.7\textwidth]{#4}}%
  \ifdim\ht\IBoxA>\ht\IBoxB
    \setlength\IHeight{\ht\IBoxB}%
  \else\setlength\IHeight{\ht\IBoxA}\fi
  \begin{figure}[!htb]
  \minipage[t]{0.45\textwidth}\centering
  \includegraphics[height=\IHeight]{#1}
  \caption{#2}\label{#3}
  \endminipage\hfill
  \minipage[t]{0.45\textwidth}\centering
  \includegraphics[height=\IHeight]{#4}
  \caption{#5}\label{#6}
  \endminipage 
  \end{figure}%
}

\begin{document}

\mainmatter  

\title*{EmbPred30: Assessing 30-days Readmission for Diabetic Patients using Categorical Embeddings}

\titlerunning{Assessing 30-days Readmission for Diabetic Patients}

\author{Sarthak\inst{1} \and
*Shikhar Shukla\inst{2} \and
*Surya Prakash Tripathi\inst{3}}
%

%
\institute{Sarthak \at Analytics Quotient, Bangalore, India \\
\email{sarthak.sfc@gmail.com, sarthak.j@aqinsights.com} \and
Shikhar Shukla\at Samsung R\&D Institute India-Bangalore, Bangalore, India\\
\email{shikhar.00778@gmail.com, shikhar.0077@samsung.com} \and
Surya Prakash Tripathi \at Institute of Engineering and Technology, Lucknow\\
\email{surya.tripathi@ietlucknow.ac.in} 
}

\maketitle

\abstract{
Hospital readmission is a crucial healthcare quality measure that helps in determining the level of quality of care that a hospital offers to a patient and has proven to be immensely expensive. It is estimated that more than \$25 billion are spent yearly due to readmission of diabetic patients in the USA. This paper benchmarks existing models and proposes a new embedding based state-of-the-art deep neural network(DNN). The model can identify whether a hospitalized diabetic patient will be readmitted within 30 days or not with an accuracy of 95.2\% and Area Under the Receiver Operating Characteristics(AUROC) of 97.4\% on data collected from 130 US hospitals between 1999-2008. The results are encouraging with patients having changes in medication while admitted having a high chance of getting readmitted. Identifying prospective patients for readmission could help the hospital systems in improving their inpatient care, thereby saving them from unnecessary expenditures. 
}


\section{Introduction}
Diabetes is a disease-causing high level of blood sugar. In type 1 Diabetes, body doesn't produce insulin, but if injected from external sources, will use it and in type 2, the body doesn't produce as well as use insulin. It is estimated that 30.3 million people of all ages in the US are suffering from Diabetes as of 2015, out of which 7.2 million are unaware\cite{US_DM}. As of 2016, it is ranked seventh in the list of global causes of mortality. Diabetes can be an underlying cause for many cardiovascular diseases, retinopathy, and nephropathy leading to frequent readmission in the hospital.

The Centers for Medicare and Medicaid Services(CMS) labeled a 30-day readmission rate as a measure of healthcare quality offered by the hospital in order to provide the best inpatient care and improve the healthcare quality. Hospitals with high readmission rates will be penalized as per the Patient Protection and Affordable Care Act(ACA) of 2010\cite{CMS}. During the recent studies\cite{Ostling}, it was observed that a 30-day readmission rate for patients with Diabetes ranges between 14.4\%-22.7\%,  which is significantly higher than the rate of all 30-day readmitted patients(8.5\%-13.5\%). In 2012, expenditure incurred due to hospital admissions amounted to \$124 billion, of which \$25 billion were spent on readmitted patients\cite{Rubin2018}. 

A popular approach of handling categorical variables in tabular data is to convert every unique value into a one-hot encoded vector of n-dimensions (where n is the cardinality of that variable). Much of researches, that have been conducted in predicting the readmission of a patient, have reduced the cardinality in a categorical feature to prevent the models from the curse of dimensionality. In this paper, we propose an approach of projecting every categorical feature in our UCI\cite{UCI} dataset into a k-dimensional vector space allowing the model to learn by itself which categories in a feature have a similar representation and which have an important role in determining 30-day readmission. 

The main purpose of this research is to facilitate healthcare institutions in predicting readmission of a diabetic patient by allowing the model to learn the relation among features and their importance in determining whether the patient will be readmitted or not. This helps the hospitals in providing the best inpatient treatment and improve the cost efficiency of healthcare centers. At the same time, it is important to identify the key factors responsible for the readmission of a diabetic patient.

The structure of the work is as follows. In Section 2, we discuss the previous related studies in this field. Section 3 deals with the motivation behind choosing the current techniques and model architecture for predicting a 30-day readmission, dataset description and data preprocessing. In Section 4, we present the experimental results. Finally, in Section 5, we discuss the conclusions drawn from the experiment and future work.

\section{Related work}

Over the years, several machine learning and deep learning models have been developed in an effort to reduce the hospital's readmission rate. LACE index (Length of stay, Acuity of admission, Charlson comorbidity index and Emergency visits) was the most preferred model due to its ease of implementation\cite{Damerye016921,DamianhbA1c,Low}. However, due to an unbalanced dataset, the models achieved low c-statistics or ROC scores ranging from 56\%-63\%. 

Munnagi et al.\cite{Munnangi} used Stepwise Regression, Forward Regression, LARS\cite{LARS} or more commonly known as Least Angle Regression, and LASSO\cite{LASSO} for feature selection and trained several models based on Logistic regression, Decision Trees, Gradient Boosting, and SVMs. His research showed that SVM outperformed all other models with an ROC of 63.3\%. 

Rubin et al.\cite{Rubin2018} proposed DERRI (Diabetes Early Readmission Risk Index), which is a multivariate logistic regression with an ROC of 72.0\%. DERRI was trained on 13 statistically significant features selected from 43 features and first proposed that HBA1C level has little relation to that of a 30-day readmission.  According to his research, lower socio-economic status, racial/ethnic minority, comorbidity burden, public insurance, emergent or urgent admission, and a history of recent prior hospitalization are some of the important factors responsible for 30-day readmission of patients.

Bhuvan et al.\cite{Bhuvan2016IdentifyingDP} trained a deep learning model having a single hidden layer containing two nodes and using mean square error as a loss metric plus a quadratic penalty with the BFGS method\cite{BFGS}. The model was able to predict readmission with an area under the precision-recall curve of 23\%. The model is trained on an unbalanced dataset and performs better than many baselined machine learning algorithms such as Logistic Regression, Naive Bayes, Adaboost trees, and Random Forest. They used an approach of using Random Forest for Feature selection which seemed to be pivotal in many researches carried out later. Their research indicated that number of inpatient visits, discharge disposition, number of laboratory tests, and admission type were important in determining readmission of a patient.

Damian et al.\cite{DamianhbA1c} also demonstrated that while HBA1c levels are important, they might not be crucial in predicting readmission of a patient. He built a different model for the age groups [0-30),[30-70) and [70-99). An ensemble model comprising of extreme gradient boosted trees, gradient boosted greedy trees and extratrees classifiers\cite{Geurts2006} designed for agegroup [0-30) achieved an accuracy of 84.8\%. For the agegroup [30-70), an ensemble model containing random forest using Gini function, gradient and extreme gradient boosted trees with early stopping achieved an accuracy of 78.5\%. For the agegroup [70-99), an ensemble model containing extratrees classifiers\cite{Geurts2006} and extreme gradient boosted trees with early stopping  achieved an average accuracy of 68.5\%. The average accuracy across three models comes out to be 77.2\%. 

Chopra et al.\cite{Chopra} used Recurrent Neural Network achieving an ROC of 80\% with 81.12\% accuracy outperforming all models based on Neural Networks, SVM, Random Forest and logistic regression on the UCI\cite{UCI} dataset developed before it. The research uses entire data of 100,000 readmissions without feature selection and feature engineering, providing much scope for improvement. A novel approach of using Approximate Bayesian Bootstrap\cite{ABB} for filling the missing data was also implemented in this research.

Ahmad et al.\cite{Hammoudeh}, in his research, designed a model using Convolutional Neural Networks (CNN), which was trained after data preprocessing, feature selection, and feature engineering. SMOTE\cite{SMOTE} was used to tackle the problem of an unbalanced dataset by generating random data of patients readmitted within 30 days. The model achieved an accuracy of 92\% with an ROC of 95\%. The research points out the need for data pruning techniques such as removing the duplicate patient data and outliers in the dataset.

Ching-Yi Lin et al.\cite{CYLIN} employs several preprocessing techniques such as data cleaning, standardization, log transformations, data balancing, and feature engineering to develop a model using logistic regression, decision trees and random forests. The research found that the number of medicines used, along with high levels of HBA1c, and discharge to another unit in the same hospital had a high correlation with the chances of a patient getting readmitted within 30 days. Random forest using the Gini function outperformed other machine learning algorithms, and achieved accuracy and ROC of 94\%.

Pham et al.\cite{Pham}, after extensive feature selection and feature engineering, proposed an ensemble model consisting of 5 models selected out of 15 models designed, which were variants of logistic regression, Decision trees, Neural networks, and Augmented Naive Bayes\cite{ANB}. These models were selected after analyzing their accuracy and their Jaccard similarity to maximize the accuracy and sensitivity of the ensemble model. The ensemble model on an unbalanced dataset achieved an accuracy of 63.5\%.

Goudjerkan et al.\cite{Jayabalan} benchmarks several existing models employing LACE, machine learning, and deep learning models, and compiles the best of every research into a single multilayer perceptron model consisting of two hidden layers with dropout\cite{srivastava2014dropout} to achieve high overall accuracy and ROC. The research makes use of comprehensive data cleaning, data reduction, and data transformation techniques. Random Forest algorithm is used for feature selection and SMOTE\cite{SMOTE} algorithm for data balancing. The model reports a k-fold cross-validation accuracy and an ROC of 95\%. 

In Table~\ref{tab:prev}, we summarize the quantitative results of the above previous studies. It includes the model basis, if the dataset had been augmented(Data Aug.) and AUROC. The results of our proposed models are also mentioned (at the top) in the table.

\begin{landscape}
\begin{table*}[]
\vspace*{-0.6cm}
\centering
\caption{Quantitative Review of Previous Studies along with our Results.}
\label{tab:prev}
\begin{tabular}{L{1cm}L{6cm}L{1.5cm}L{1.5cm}L{7.6cm}L{0.8cm}}
\toprule
\textbf{Year} & \textbf{Model basis} & \textbf{Data Aug.} & \textbf{AUROC} & \textbf{Remarks} & \textbf{Ref.} \\ 
        \midrule
        2019 & Categorical Embeddings and Neural Networks & Yes & \textbf{97.4} & Embeddings for categorical features concatenated with normalised continuous features were fed into neural network.  &  self \\
        \midrule
        2019 & Multilayer Perceptron & Yes & 95.0 & Extensive data augmentation and feature engineering was used to train a multilayer perceptron network. &  \cite{Jayabalan} \\
        \midrule
        2019 & Ensemble of Logistic Regression,Decision Tree,Neural Network and Augmented Naive Bayes Network & No & 63.5$^1$ & Ensemble of models were used to improve model AUROC and sensitivity in an unbalanced dataset. &  \cite{Pham} \\
        \midrule
        2018 & Random Forest & No & 94.0 & After Feature Selection and Feature engineering, Random Forest model was trained using Gini function.  &  \cite{CYLIN} \\
        \midrule
        2018 & Convolutional Neural Network & Yes & 95.0 & Feature Selection and Feature engineering coupled with convolutional and linear layers was used. &  \cite{Hammoudeh} \\
        \midrule
        2017 & Recurrent Neural Network & No & 81.1 & Proposed a Recurrent Neural Network(RNN) that performed better than Logistic Regression, SVMs, Multilayer Perceptrons on unbalanced dataset. &  \cite{Chopra} \\
        \midrule 
        2016 & Ensemble model using Extreme Gradient Boosted Trees with early stopping, SVM, RandomForest and Extratrees classfier & No & 77.2$^1$ & Built three different ensemble machine learning models for three age groups,0-30,30-70 and 70-100. &  \cite{DamianhbA1c} \\
         \midrule
        2016 & Neural Networks & No & 23.0$^2$ & A single hidden layer was used for prediction based on features selected and features engineered. &  \cite{Bhuvan2016IdentifyingDP} \\
        \midrule
        2015 & Multivariate Logistic Regression & No & 72.0 & Feature Selection and Feature engineering was used to train a logistic regression model. &  \cite{Rubin2018} \\
        \midrule
        2015 & SVM & No & 63.3 & Optimal parameters estimated using LARS\cite{LARS} were using in training an SVM. &  \cite{Munnangi} \\
 \bottomrule
 \multicolumn{6}{l}{$[1]$ - Accuracy; 
 $[2]$ - Area under the precision-recall curve} \\ 
 \multicolumn{6}{l}{*All the results are on UCI\cite{UCI} dataset}. 
\end{tabular}%
\end{table*}
\end{landscape}

\section{Proposed Method}

\subsection{Motivation}

Several state-of-the-art results\cite{Jayabalan,Hammoudeh} of predicting hospital readmission using neural networks have been obtained using data augmentation using SMOTE\cite{SMOTE} for unbalanced classes. Combining neural networks with researches\cite{CYLIN,Rubin2018,Hammoudeh,Chopra,Jayabalan} that performed extensive feature engineering and feature selection seemed to enhance the performance. This motivated us to improve the existing methods of feature selection and feature engineering while using neural networks.

Recently, FastAI\cite{howard2018fastai} proposed a tabular data model for structured data, having categorical and continuous features. It combines several state-of-the-art methods such as creating embeddings of optimal sizes for categorical data for training an efficient model and extracting essential features from the structured data. 

In our research, we augment existing data using SMOTE\cite{SMOTE} to handle unbalanced classes, generate new features and train an efficient model using FastAI\cite{howard2018fastai} tabular model.

\subsection{Dataset}

This research uses the dataset obtained from the Center for  Machine Learning and Intelligent Systems at the University of  California, Irvine\cite{UCI} covering 10 years(1998-2008) of Diabetes patient data gathered from 130 US hospitals having 70,000 distinct patients\cite{Strack} with over 100,000 records. Every record was labeled as to whether the patient was readmitted within 30 days, readmitted after 30 days, or not readmitted at all.

\subsection{Data Preprocessing}

\renewcommand{\theenumi}{\roman{enumi}}%
\begin{enumerate}
    \item \textbf{Filling missing values}: The missing values in the categorical columns are replaced by ‘nan’ which is treated as another category. On the other hand, for filling the missing data in continuous-valued columns, we experimented with three commonly used techniques. Replacing missing data with:
    \begin{itemize}
        \item Average of the column 
        \item Median of the column
        \item Constant value of ‘0’
    \end{itemize}
    Out of these three, the best results were obtained by filling with a median of the column since it is the best representative of the distribution of data.
    
    \item \textbf{Removing Outliers and Inconsistencies}:  As suggested in a recent research\cite{CYLIN}, it is important to maintain a single record for every patient id. So continuing along the path of this research, only the first encounter with the patient is kept, and the rest of the patient's records are dropped. 
    
    Also, as suggested by the dataset\cite{UCI}, the data contains records of those patients who have diagnosed with Diabetes, in at least one of the three diagnoses done by hospital staff. If none of the columns \emph{diag\_1}, \emph{diag\_2}, \emph{diag\_3} (which represent three different diagnoses done by the hospital) did not have the Diabetes ICD9 code mentioned, which is of the format \emph{250.xx}, then those records were also dropped. 
    
    Also, if a patient died or is referred to the end of life care, then those patients would not be readmitted and hence were deleted from the dataset. 
    
    \item \textbf{Feature Encoding}: The dataset contains three classes, with 11.2\% of the 70,000 patients readmitted within 30 days, 34.9\% readmitted after 30 days, and 53.9\% are not readmitted at all. Since we have to predict whether a patient will be readmitted within 30 days or not, dropping patients readmitted after 30 days would result in the loss of one-third of data. Therefore, we define two classes, \emph{Yes} and \emph{No}, \emph{Yes} suggesting that the patient will be readmitted within 30 days else \emph{No}.
    
    Levels of HBA1C and Glucose serum test results are also relabeled into three categories, namely “normal”(Normal), “abnormal”(Values $>$7,$>$8 for HB1Ac, and $>$300,$>$200 for Glucose serum test) and “not tested”(None). 
    
    All the 23 columns related to the amount of medicines administered to an admitted patient are also relabeled into two categories, \emph{Yes} or \emph{No}. If the level of any medicine is \enquote{Steady} or \enquote{None} then it is labeled as No, and Yes if the level is \enquote{Up} or \enquote{Down}.

    \item \textbf{Data Synthesis}: To deal with the imbalance in the original dataset, which has 88.8\% negative samples, we generated data through SMOTE (Synthetic Minority Oversampling Technique)\cite{SMOTE}. This statistical technique takes neighbouring samples from the minority target class and generates new samples whose features are a combination of the original samples’ feature space. After using this technique our synthetic dataset has a 1:1 ratio of samples from both the target classes.

    \item \textbf{Normalizing continuous variables}: Many columns such as \enquote{number\_inpatient},\enquote{number\_outpatient},\enquote{number\_emergency} were highly skewed with high kurtosis. To reduce the skewness, log(x+1) transformation was used. The features in the dataset vary in scale, unit, and range. Due to this, the features with a broad range in the dataset can have a disproportionate impact on the prediction. To overcome this problem, the data is normalized so that each feature has a mean of 0 and a standard deviation of 1.
    
    \item \textbf{Feature Engineering}:
        \begin{itemize}
            \item \emph{Feature Synthesis}:  In the previous researches\cite{Jayabalan,Hammoudeh,CYLIN}, two new features were created \emph{service\_utilization} and \emph{count\_meds}, which proved influential in determining whether the patient will be readmitted within 30 days or not. 
            
            \emph{Service utilization} is the sum of the number of times a patient has used the services of the hospital. This is calculated by summing up the number of inpatient visits, the number of outpatient visits, and the number of emergency visits.
            
            \emph{Count\_Meds} is a feature counting the number of medication changes that happened when a patient was admitted. This was done by counting the number of \enquote{Yes} in the 23 columns related to the amount of medicine registered.
            \item \emph{Feature Selection}: Out of 49 columns present in the dataset, columns that had more than 50\% values missing, such as weight, were dropped. Two columns \emph{examide} and \emph{citoglipton} had their cardinality as 1; the only value being \emph{No}. As a result, these two columns were also dropped.
            
            The importance of the rest of the columns was analyzed after training our model for 70 epochs. Most of the columns or features which had importance of 0 were dropped. One of the features, \emph{Count\_Meds}, which was engineered, was determined as the most important feature among all others. In Fig.~\ref{fig:feature importance} we present the importance of final features fed into our model.
            
            \begin{figure}
            \centering
            \resizebox{1.0\textwidth}{!}{%
            \includegraphics[width=\textwidth]{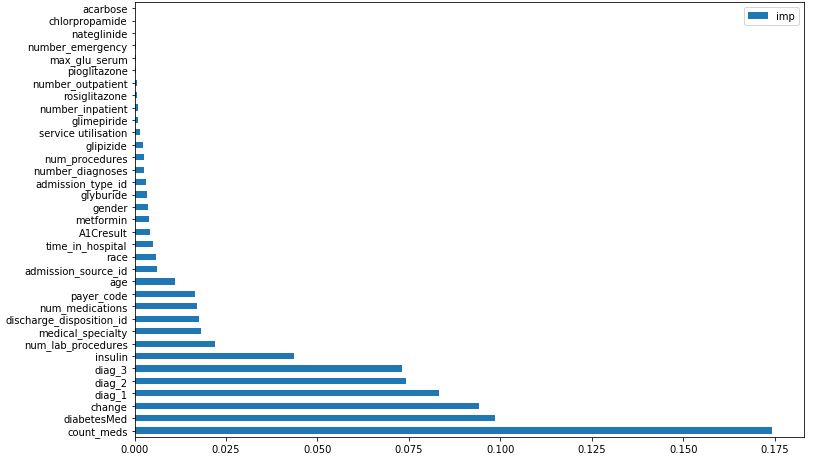}
            }
             \caption {Input features of our model}
             \label{fig:feature importance}
            \end{figure}
        \end{itemize}
\end{enumerate}

\subsection{Model Details}

Categorical variables are passed through an embedding layer. This layer converts each categorical value to a float vector with a dimension given by the formula\cite{howard2018fastai}: 
\begin{equation}
	\mathrm{Embedding \ dimension} = [1.6 * n^{0.56}],
\end{equation}
where n is the cardinality of the column.

\begin{table}[]
\centering
 \setlength{\belowcaptionskip}{-5pt}
 \caption{Architecture of the Embedding Model}
 
 \label{tab:arch_model}
 \scalebox{0.8}{
\resizebox{\textwidth}{!}{%
\begin{tabular}{|c|c|c|c|}
\hline
\multirow{2}{*}{Layer Name} & \multirow{2}{*}{Output features} & Momentum / & No. of \\                  
&  & eps(x-epsilon) & parameters \\\hline
(Embedding Layer) & & & \\
race                & (1,4)     &  & 24 \\
gender              & (1,3)     &  & 9  \\
age                 & (1,6)    &  & 66  \\
payer\_code          & (1,8)    &  & 144 \\
medical\_specialty   & (1,17)   &  & 1,207  \\
diag\_1              & (1,62)  &  & 42,532 \\
diag\_2              & (1,63)  &  & 43,911 \\
diag\_3              & (1,65)  &  & 47,840  \\
max\_glu\_serum       & (1,3)     &  & 12  \\
A1Cresult           & (1,3)     &  & 12 \\
metformin           &  (1,3)     &  & 9  \\
nateglinide         &  (1,3)     &  & 9  \\
chlorpropamide      & (1,3)     &  & 9  \\
glimepiride         & (1,3)     &  & 9  \\
pioglitazone        & (1,3)     &  & 9  \\
rosiglitazone       & (1,3)     &  & 9  \\
acarbose            & (1,3)     &  & 9  \\
insulin             & (1,3)     &  & 9  \\
change              & (1,3)     &  & 9  \\
DiabetesMed         & (1,3)     &  & 9  \\
\hline
(Embedding Dropout) &   &   &  \\
Dropout(0.05)       &  (283) &   & 283 \\
\hline
(Batch Normalization & & &\\
for continuous variables) & & & \\
BatchNorm1d  & (13) & 0.1 / 1e-05 & 13 \\
\hline
(Sequential) & & & \\ 
Linear & (512) & 0.1 / 1e-05  & 144,896 \\
RELU & (512) & & 512 \\
BatchNorm1d & (512) & 0.1 / 1e-05 & 512 \\
Dropout(0.15) & (512) & & 512 \\
Linear & (512) & 0.1 / 1e-05 & 262,144 \\
RELU & (512) & & 512 \\
BatchNorm1d & (512) & 0.1 / 1e-05 & 512 \\
Dropout(0.15) & (512) & & 512\\
Linear & (2) &  & 1,024\\
\hline
Total Parameters & & & 547,279\\
\hline
\end{tabular}%
}
}
\end{table}

For each categorical column, the embedding matrix has rows corresponding to unique categorical values in the dataset. After this, dropout is applied to the values obtained from the previous operation. The continuous variables are passed through a batch normalization layer. The results from both the continuous and categorical variables are then concatenated and fed to a feedforward network whose architecture is described in Table \ref{tab:arch_model}.

\section{Results and Discussion}

\begin{figure}
        \hspace*{-0.5cm}
        \setbox0\hbox{%
                \includegraphics[width=0.62\textwidth]{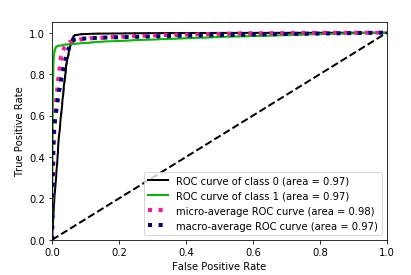}%
        }%
        \setbox2\hbox{%
                \includegraphics[width=.62\textwidth]{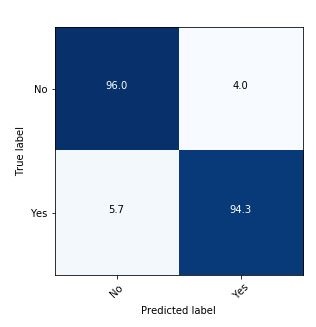}%
        }%
        \ifdim\ht0>\ht2
                \setbox0\hbox{%
                        \includegraphics[height=\ht2]{roc.jpg}%
                }%
        \else
                \setbox2\hbox{%
                        \includegraphics[height=\ht0]{confusion_matrix.jpg}%
                }%
        \fi
        \noindent
        \parbox{.55\textwidth}{%
                \centering
                \unhbox0
                \caption{ROC Curve}
                \label{fg:ROC}
        }%
        \hfil
        \parbox{.45\textwidth}{%
                \centering
                \unhbox2
                \hspace*{1.5cm}\caption{Confusion Matrix}
                \label{fg:confusion}
        }%
\end{figure}

This paper's approach achieves state-of-the-art result on the UCI dataset\cite{UCI} in determining whether a patient will be readmitted within 30 days or not. In Fig.~\ref{fg:ROC}, we plot the roc curve for two classes in our dataset and in Fig.~\ref{fg:confusion}, we present the confusion matrix of our trained model. Overall the model's accuracy is \textbf{95.2\%} with a standard deviation of \textbf{0.34\%} and ROC is \textbf{97.4\%} with a standard deviation of \textbf{0.42\%} after evaluating it using k-fold cross-validation(k=6).

\subsubsection{Future Scope} The original dataset was severely imbalanced; the availability of data with more samples of the underrepresented class could help mitigate this problem. This would provide more realistic data points than the ones generated by SMOTE\cite{SMOTE} algorithm. Also, there can be experimentation with other machine algorithms to gauge their performance. Furthermore, there can be more analysis of the embedding matrices to help interpret and visualize the distinguishing features. The effect of varying the dimensions of each embedding is another potential area of study.

\section{Conclusion}

This research analyzed the readmission patterns of diabetic patients using unsupervised methods and proposes a novel approach of generating embeddings for categorical features such as those of ICD9 codes used while diagnosing the patient. We performed various feature engineering and feature selection strategies along with data augmentation for keeping the issue of unbalanced classes at bay. 

To exceed the ROC score of the best existing model, we combined our approach with the best of previous researches to obtain the state-of-the-art ROC of \textbf{97.4\%} and an accuracy of \textbf{95.2\%}. We found that changes in the number of medication administered to a patient play a vital role in determining whether a patient will be readmitted to a hospital within 30 days or not.

\bibliographystyle{acm}
\bibliography{References}

\end{document}